\documentclass[journal]{IEEEtran}
\ifCLASSINFOpdf
  \usepackage[pdftex]{graphicx}
\else
\fi

\usepackage{xcolor}
\usepackage{multirow}
\usepackage{balance}

\begin{document}
%
\title{Knowledge Learning with Crowdsourcing: A Brief Review and Systematic Perspective}
%
%
%

\author{Jing~Zhang,~\IEEEmembership{Senior Member,~IEEE}
\thanks{Manuscript received August 22, 2021; revised November 5, 2021; accepted December 10, 2021. This work was supported by the National Key Research and Development Program of China (2018AAA0102002), and the National Natural Science Foundation of China (62076130, 91846104). Recommended by Associate Editor Long Chen.}
\thanks{Citation: J. Zhang, “ Knowledge learning with crowdsourcing: A brief review and systematic perspective,” IEEE/CAA J. Autom. Sinica, vol. 9, no. 5, pp. 749–762, May 2022. Digital Object Identifier 10.1109/JAS.2022.105434}
\thanks{J. Zhang was with the School of Computer Science and Engineering, Nanjing University of Science and Technology, Nanjing 210094, China. E-mail: jzhang@njust.edu.cn.}
\thanks{This version will be continually updated according to the recent research progress in this field.}
}

%
%

\markboth{IEEE/CAA Journal of Automatica Sinica,~Vol.~9, No.~5, May~2022}%
{Zhang: Knowledge Learning with Crowdsourcing}
%



\maketitle

\begin{abstract}
Big data have the characteristics of enormous volume, high velocity, diversity, value-sparsity, and uncertainty, which lead the knowledge learning from them full of challenges. With the emergence of crowdsourcing, versatile information can be obtained on-demand so that the wisdom of crowds is easily involved to facilitate the knowledge learning process. During the past thirteen years, researchers in the AI community made great efforts to remove the obstacles in the field of learning from crowds. This concentrated survey paper comprehensively reviews the technical progress in crowdsourcing learning from a systematic perspective that includes three dimensions of data, models, and learning processes. In addition to reviewing existing important work, the paper places a particular emphasis on providing some promising blueprints on each dimension as well as discussing the lessons learned from our past research work, which will light up the way for new researchers and encourage them to pursue new contributions.
\end{abstract}


%
\IEEEpeerreviewmaketitle

\section{Introduction}
%
%
%
%
\IEEEPARstart{I}{n} today's era of big data, the acquisition of massive raw data is no longer a tricky thing, but exploring and exploiting knowledge from these data is still full of challenges. Facing the enormous volume, high velocity, diversity, value-sparsity, and uncertainty of big data, the knowledge learning process has never been entirely automated, which though is a beautiful vision in the artificial intelligence (AI) research community. For example, many deep learning models that have achieved great successes in recent years still heavily rely on large datasets with good annotations provided by skilled humans. Therefore, current knowledge discovery and learning process are still inseparable from the investment of a large amount of human labor and wisdom. The emergence of crowdsourcing has provided a viable solution to this difficulty. Crowdsourcing is defined as the practice of obtaining information or input into a task or project by enlisting the services of a large number of people, either paid or unpaid, typically via the Internet \cite{Howe2006rise}. Instead of seeking domain experts to perform data pre-processing, requesters are looking for workers entirely from the Internet to handle raw data, among whom there are professionals, ordinary people, spammers, and even some adversaries. Compared with the traditional way of hiring experts, resorting to crowdsourcing is faster, lower-cost, more creative, but also with more unneglectable uncertainty. Although imperfect, learning with crowdsourcing has appeared many successful cases, from natural language processing \cite{sabou2012crowdsourcing}, computer vision \cite{kovashka2016crowdsourcing}, bioinformatics \cite{good2013crowdsourcing} to medical diagnosis \cite{li2017reliable}. Crowdsourcing has created many opportunities for AI-related disciplines \cite{Vaughan2017MakingBU}.

AI Community first foresaw the opportunities brought by crowdsourcing to knowledge discovery and machine learning around 2008. In 2008, there appeared two milestone studies \cite{sheng2008get,Snow2008Cheap}. Sheng \textit{et al.} \cite{sheng2008get} investigated the \textit{repeated labeling} and majority voting scheme, where requesters ask multiple crowd workers to label the same objects and then determine the (integrated) labels of the objects by voting the different judgments. In addition, they also investigated the impact of integrated labels on the learned prediction models. Compared with \cite{sheng2008get} that focused on general learning problems by simulation, Snow \textit{et al.} \cite{Snow2008Cheap} specified their research on five natural language processing tasks, collecting annotations from the Amazon Mechanical Turk (MTurk) crowdsourcing platform. They used a classical Dawid \& Schene's model \cite{dawid1979maximum} to integrate multiple noisy labels and showed that the quality of integrated labels can meet the natural language processing (NLP) requirements if good label aggregation algorithms are used. This work showed the usability of real crowdsourcing annotations in knowledge learning tasks for the first time. From then, during the past thirteen years, researchers in the AI community have developed many techniques to tackle the defects of using crowdsourcing in machine learning. A large number of studies focused on general-purpose technologies of learning from crowdsourced annotated data \cite{zhang2016learning}, including statistical truth inference \cite{zheng2017truth}, predictive model training with noisy labels \cite{sheng2008get,raykar2010learning,Kajino2012Convex,bi2014learning}, optimization for cost-effectiveness trade-off \cite{welinder2010online,Rokicki2014CompetitiveGD,huang2017cost}, etc.

\begin{figure*}
	\centering
	\includegraphics[width=6.8in]{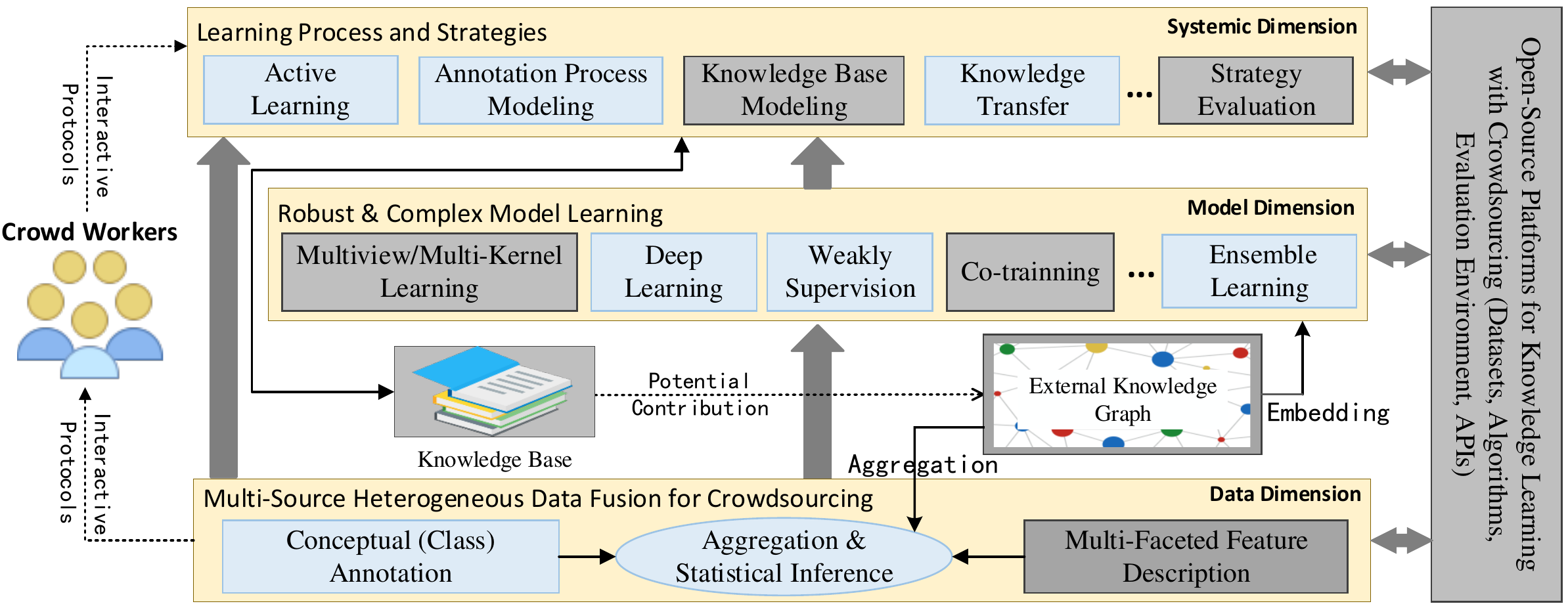}
	\caption{Our systematic perspective of a knowledge-learning-with-crowdsourcing framework. The gray parts are what currently has not been well studied (only a few or no studies can be found). The interactive protocols between crowd workers and knowledge learning systems are out of the scope of this paper.}\label{figArch}
\end{figure*}

There are several articles reviewing the technical progress in this field from different angles. For example, Daniel \textit{et al.} \cite{daniel2018quality} reviewed the literature from the perspective of quality control for crowdsourcing. Besides quality control, Chittilappilly \textit{et al.} \cite{chittilappilly2016survey} comprehensively reviewed a wider work including incentive design and task assignment. Another previous survey \cite{zhang2016learning} focused on both truth inference and predictive model learning while article \cite{zheng2017truth} merely focused on truth inference. However, several years have passed since these reviews were published, and more new technologies have emerged recently. Compared with \cite{Sheng2019Machine} that reviews the studies from the machine learning viewpoint and also serves as a basis of this study, this paper proposes a bigger systematic knowledge learning framework and derives many interesting and meaningful research topics that existing studies have not touched or deeply investigated.

The objectives of this paper are as follows: 1) It briefly reviews the typical general-purpose technology in this field from the perspective of knowledge learning in the knowledge discovery from data (KDD) process, which will help readers quickly understand the main scientific issues in the field of knowledge learning with crowdsourcing, the development context of mainstream technologies, and the current state-of-the-art achieved; 2) It also provides our viewpoints on the development direction of this field, which may illuminate young researchers who want to enter this field. As a concentrated review paper with our perspectives, we do not intend to include every research work with trivial contributions. Instead, we particularly emphasize the forecast of the development trend of techniques and the construction of a larger systematic blueprint that encompasses these techniques. More precisely, what makes this paper quite different from the previous survey papers, which is also its contributions, lies in three points:

\begin{enumerate}
\renewcommand{\labelenumi}{(\theenumi)}
	\item We embrace existing techniques into a systematic framework from the perspective of knowledge learning. Differently from the previous machine learning-oriented thought, we believe that the improvement of the knowledge learning process can reflect the advantages of crowdsourcing in terms of diversity. Besides, a new trend in machine learning is to add domain knowledge into learned models. Therefore, a knowledge learning-oriented framework is more accordant with the technical development trend.
	\item We are not intended to comprehensively review all existing studies. Instead, we more emphasize our own thoughts while reviewing the typical progress. We have obtained some experience and lessons in the past ten years of research in this field, which propels us to propose a knowledge-learning framework to deal with future challenges.
	\item We present many future research topics in different dimensions of our knowledge-learning framework and also discuss the technical roadmaps to realize them as well as provide some primary ideas of the solutions, which makes this paper not only a mirror reflecting the past but a guide to the future.
\end{enumerate}

Fig.~\ref{figArch} shows our proposed systematic perspective of a knowledge-learning framework for crowdsourcing. We categorize the techniques into three dimensions\footnote{Although Fig.~\ref{figArch} appears a hierarchical structure, we use the term \textit{dimension} while not \textit{layer} because these techniques are independent from one another and have no strict interfaces between them. Thus, they do not form a hierarchical structure.}: data, model, and systemic. Each dimension has its own research contents and objectives. The lower-layer techniques can provide supports for the upper layers or directly be used by the upper layers. The data dimension focuses on data fusion from multiple heterogeneous crowdsourced sources (workers and raw data). It provides various kinds of data with different qualities for the model dimension. The model dimension uses these (maybe noisy) data to train robust and complex predictive (knowledge) models.  The systemic dimension seeks the techniques that can optimize the knowledge learning process, including reducing the cost, enhancing the capability of workers, and improving the availability of knowledge learning systems. All of them form an overall solution for knowledge learning with crowdsourcing. Now, we discuss the techniques in each dimension and their development trends.

\begin{table*}[]\centering
	\caption{Taxonomy for Agnostic True Inference Methods for Crowdsourced Annotation}
	\label{tab:inference}
	\begin{tabular}{|c|c|c|c|c|}
		\hline
		Category       & Binary-class & Multi-class & Multi-label & Numeric \\ \hline
		Generative     & \begin{tabular}[c]{@{}c@{}}RY \cite{raykar2010learning}, GLAD \cite{whitehill2009whose},\\ \cite{bi2014learning}, \cite{kurve2015multicategory}, \cite{welinder2010online}, \\ IV-BP/MF \cite{liu2012variational}, CBias \cite{Gemalmaz2021Accounting} \end{tabular} 
		&\begin{tabular}[c]{@{}c@{}}DS \cite{dawid1979maximum}, Multi-D \cite{welinder2010multidimensional}, BCC \cite{kim2012bayesian},\\ ZenCrowd \cite{demartini2012zencrowd}, LC-ME \cite{tian2015uncovering}, \\ MinimaxEntropy \cite{zhou2012learning}, cBCC \cite{Venanzi2014Community},\\ SpectralDS \cite{zhang2016spectral}, EBCC \cite{li2019Exploiting}, \\ \cite{zhou2014aggregating}, BayesDGC \cite{Li2021CrowdsourcingAW} \end{tabular}
		&\begin{tabular}[c]{@{}c@{}} MLNB \cite{bragg2013crowdsourcing}, P-DS \cite{duan2014separate}, \\ ND-DS \cite{duan2014separate}, MCMLD \cite{zhang2018multi},\\ MCMLD-OC \cite{Zhang2021MultiLabelTI} \end{tabular} 
		& RY\_N \cite{raykar2010learning} \\ \hline
		Discriminative &\begin{tabular}[c]{@{}c@{}}MV, \cite{jung2011improving}, KOS \cite{karger2011budget},\\ \cite{ghosh2011moderates}, \cite{dalvi2013aggregating}, PLAT \cite{zhang2015imbalanced}, \\ IEThresh \cite{donmez2009efficiently} \end{tabular}
		&\begin{tabular}[c]{@{}c@{}}PV, \cite{aydin2014crowdsourcing,Tao2020LabelSW}, CATD \cite{Li2014ACA}, PM\cite{Li2014ResolvingCI},\\ \cite{tian2015max,zhou2016crowdsourcing}, MNLDP \cite{Jiang2021LearningFC}, \\ GTIC \cite{zhang2016multi}, \cite{gaunt2016training}, LLA \cite{yin2017aggregating}, \\ CrowdLayer \cite{rodrigues2018deep}, SpeeLFC \cite{Chen2020StructuredPE} \end{tabular}
		& MLCC \cite{Tu2020MultilabelCC}
		&\begin{tabular}[c]{@{}c@{}}Mean, Median \\ CATD\_N \cite{Li2014ACA}, PM\_N\cite{Li2014ResolvingCI}\end{tabular}\\ \hline
	\end{tabular}
\end{table*}

\section{Data Fusion for Crowdsourcing}
Data acquisition is one of the basic goals for us to use crowdsourcing. Since the data are provided by crowd workers with different characteristics, together with the raw data published on the platforms, this process can be viewed as a kind of data fusion from multiple heterogeneous sources. The machine learning and data mining community first realized the opportunity that crowdsourcing brought to supervised learning, i.e., obtaining class labels for training sets. To improve the quality of labels, both Sheng \textit{et al.} \cite{sheng2008get} and Snow \textit{et al.} \cite{Snow2008Cheap} proposed a \textit{repeated-labeling} scheme in 2008, which let multiple crowd workers to label the same objects and the true labels of the objects are inferred from these multiple noisy labels. Using the repeated-labeling scheme, truth inference became one of the fundamental topics in knowledge learning with crowdsourcing. In crowdsourcing learning, truth inference is defined as a process that infers (or discovers) true values for unknown and latent variables (such as labels of instances, the community of workers, etc.) and parameters (such as difficulties of instances, reliability of workers, etc.) of crowdsourced annotation systems from the observed data including the original instances and the noisy annotations provided by the crowd workers.

\subsection{Truth Inference}
The general-purpose truth inference methods for crowdsourced annotation systems have been well studied in the past. The current mainstream statistical-based methods often work in an agnostic manner, where only observed noisy labels and the original instances are used for inference. The core function of this process is to estimate true labels for instances from their multiple noisy labels. Because the inferred labels for the instances aggregate the judgments of different crowd workers, the process is called \textit{label aggregation} or \textit{label integration}. Meanwhile, some other information, such as the reliability, dedication, and intention of workers, the difficulty of instances, etc., might be simultaneously inferred, which depends on models. Thus, truth inference provides essential knowledge about the crowdsourced annotation environment. Table~\ref{tab:inference} summarizes the agnostic true inference methods for different crowdsourced annotation tasks in this paper.

\subsubsection{Agnostic Probabilistic Methods}
A large number of agnostic inference methods are based on probability and statistics. These methods can be divided into two main categories according to their differences in probabilistic modeling: \textit{probabilistic generative} approaches and \textit{non-probabilistic generative} (\textit{discriminative}) approaches. Generative approaches use some basic probability distributions to represent the generation process of the crowdsourced labels as probabilistic graphical models, while discriminative approaches do not exactly rely on probabilistic graphical models although they can also use probability modeling.

\textit{Probabilistic generative} approaches originated from the classic Dawid \& Skene’s (DS) model \cite{dawid1979maximum}, which uses confusion matrices to model the capability of workers. DS is only based on the categorical distribution. An element $\pi_{kl}^{j}$ in a confusion matrix represents the probability of worker $j$ classifying an instance as class $l$ given its true class is $k$. Generally, DS is simple, robust and has a good explanation to the capability of a worker over each class. Raykar \textit{et al.} \cite{raykar2010learning} developed a Bayesian version of this model (namely, RY), which concentrates on modeling the workers' biases towards the positive and negative classes in binary classification tasks using two parameters \textit{sensitivity} and \textit{specificity}. RY has good performance in binary-class inference. When it is extended to multi-class tasks, the explanation to sensitivity and specificity becomes ambiguous and the model's performance will deteriorate. However, RY has a wide applicability, and it can also be extended to numeric annotations. Welinder and Perona \cite{welinder2010online} modeled the probability of a worker providing correct labels. Various aspects of an annotation process can be modeled by a generative model as long as it has a sound probabilistic explanation and generation mode. In addition to the characteristics of workers, difficulty of instances is the most common one to be modeled \cite{donmez2009efficiently,whitehill2009whose,welinder2010multidimensional,ghosh2011moderates,zhou2012learning,zhou2014aggregating}. For example, Whitehill \textit{et al.} \cite{whitehill2009whose} introduced a parameter to model the difficulty of tasks in their GLAD method. GLAD uses a logistic regression model, which makes its performance on real data sets unsatisfactory, because noise labels rarely obey a specific distribution. Welinder \textit{et al.} \cite{welinder2010multidimensional} proposed a more complicated multi-dimensional model, where noises in the instance features are also considered. More elaborate worker models can be found in \cite{bi2014learning} and \cite{kurve2015multicategory}, where the dedication and intention of workers are added into their truth inference models. Exploring and exploiting the relationship among workers brings another opportunity to improve the accuracy of inference. Based on the Bayesian classifier combination model (BCC) \cite{kim2012bayesian}, cBCC \cite{Venanzi2014Community} explores the community structure hidden behind workers by analyzing their crowdsourced labels, where workers in the same community have similar labeling results, and EBCC \cite{li2019Exploiting} exploits worker correlation to improve label aggregation. However, these models usually includes many variables and parameters, which can hardly be applied in big data environments.

Probabilistic generative models usually assume some prior distributions of variables. They have a good theoretical basis and can be solved by standard solutions for probabilistic graphical models, such as the Markov Chain Monte Carlo (or Gibbs) sampling, EM algorithms, convex optimization, and variational inference. However, their drawback lies in that if actual distributions of variables do not obey the assumptions, the inference accuracy will deteriorate. For example, the more complicated DS \cite{dawid1979maximum} and RY \cite{raykar2010learning} with predetermined assumptions occasionally perform worse than the simpler model ZenCrowd \cite{demartini2012zencrowd} (where the reliability of a worker is modeled by a binary variable) on quite a few real-world datasets \cite{zhang2016multi}. Optimization methods for objective functions in probabilistic models also affect their performance. Thus, some studies \cite{tian2015uncovering,zhang2016spectral} focus on the optimization of the models. For example, SpectralDS \cite{zhang2016spectral} uses the spectral method to obtain an initial estimates of parameters, resulting in a better outcome of its EM procedure.

\textit{Discriminative} approaches do not require that variables of the models must obey specific probability distributions and the final results are derived from a series of probabilistic inferences. Different mathematical methods such as matrix factorization and convex optimization can be used to obtain the results. The simplest discriminative method is majority voting (MV), which is also called plurality voting (PV) in multi-class cases. Although MV is simple, it is very effective. Thus, researchers are still keen to study its variants \cite{aydin2014crowdsourcing,jung2011improving,Tao2020LabelSW}. For example, Tao \textit{et al.} \cite{Tao2020LabelSW} proposed four strategies to model the similarity of crowdsourced labels. Their method gives workers different label quality weights for different samples, and finally integrates labels through weighted MV. The KOS method \cite{karger2011budget} incorporates singular value decomposition (SVD) of a low-rank matrix with a belief propagation-like procedure to achieve inference. KOS works well when the noisy label matrix is full (each worker labels all instances). Dalvi \textit{et al.} \cite{dalvi2013aggregating} proposed a similar SVD-based method, which relaxes the prerequisite that label matrix must be full. Liu \textit{et al.} \cite{liu2012variational} unified MV and KOS under a Bayesian framework (considering prior probabilities) and solved them via variational inference. Inspired by the multi-class support vector machines (SVM), Tian and Zhu \cite{tian2015max} proposed a max-margin majority voting that directly finds the most likely labels for instances by maximizing margins. Zhou and He \cite{zhou2016crowdsourcing} proposed two structured methods based on tensor augmentation and completion. The two methods use tensor representation for the labeled data, augment it with a ground truth layer, and estimate the true labels via low rank tensor completion. Jiang \textit{et al.} \cite{Jiang2021LearningFC} proposed a multiple noisy label distribution propagation method (MNLDP), which considers the relationship between multiple noisy label sets. MNLDP first estimates the distribution of noisy labels for each instance and propagates it to its nearest neighbors.  Each instance considers the noise label distribution of itself and its nearest neighbors in label aggregation. Some discriminative methods such as CATD \cite{Li2014ACA} and PM \cite{Li2014ResolvingCI} can be extended to numeric labels. Discriminative approaches are usually faster than the probabilistic generative ones.

\subsubsection{Difficulties in Agnostic Methods}
Both probabilistic generative and discriminative methods work well when label noises are regularly distributed in different categories. In reality, this premise is not always true. As early as 2013, we found that workers' labeling qualities on the two classes in binary-labeling tasks exhibit significant difference, which is the so-called \textit{biased labeling} phenomenon and will deteriorate most truth inference algorithms \cite{zhang2013imbalanced}. This cognitive bias was further confirmed by successive studies \cite{Eickhoff2018CognitiveBI,Barbera2020CrowdsourcingTT}. To deal with the bias, we proposed a PLAT \cite{zhang2015imbalanced} algorithm that can automatically adjust the decision threshold between the inferred positive and negative instances. This topic has been continually attracting the attention of researchers \cite{kamar2015identifying,Barbosa2019RehumanizedCA}. Recently, Gemalmaz and Yin \cite{Gemalmaz2021Accounting} studied a specific cognitive bias, i.e., confirmation bias, which is people’s tendency to favor information that confirms their existing beliefs and values. They proposed a probabilistic graphical model that uses a parameter to model the probability of a worker being subject to confirmation bias. Another difficulty is that if there are spammers or adversarial workers in the system, agnostic methods seldom obtain good results. Some researchers had to use the method of injecting prior information to identify low-quality workers so that their weights can be reduced during inference. For example, ELICE \cite{khattak2011quality} optimized truth inference by injecting expert labels. Not only true labels can be injected but also does the information about workers. Bonald and Combes \cite{bonald2017minimax} showed that if the reliability of a small portion of workers can be known, the reliability of all workers can be accurately inferred, and the lower bound on the minimax estimation error can be calculated. Oyama \textit{et al.} \cite{oyama2013accurate} required workers to provide their confidence levels when performing tasks. In their work, the confidence scores are utilized during inference. Liu \textit{et al.} \cite{liu2017improving} proposed a method that selects the most informative instances and maximizes the influence of expert labels injected. The method develops a complete uncertainty assessment for instance selection. The expert labels are propagated to similar instances via regularized Bayesian inference. However, these methods partially break the agnostic nature of truth inference, requiring more information and human interventions.

\subsection{Improving Aggregation with Learning Models}
It might be unwise that the above agnostic truth inference methods completely ignore the instances themselves during inference. Instance features should be helpful for obtaining better results, but how to utilize them in a domain-independent way is challenging. Some of the past research attempted to achieve the goal by using learning models in both unsupervised and supervised manners.

Zhang \textit{et al.} \cite{zhang2016multi} proposed an agnostic inference algorithm GTIC, which generates \textit{conceptual} features for instances from the crowdsourced labels of instances and uses a K-means algorithm to cluster all the instances into $K$ classes. GTIC fuzzifies the biases that are difficult to describe in multi-class classification and uses clustering to discover these biases. Based on GTIC, its subsequent method \cite{zhang2017label} also runs a clustering algorithm on the \textit{physical} features of instances and uses the clustering results to correct the potential errors in the inferred results of GTIC. In contrast, the AVNC method \cite{zhang2018improving} first builds a predictive model using a subset of the inferred data in which the instances with highly-probable wrong labels have been filtered out, and then uses this learned predictive model to correct the errors in the inferred labels. Wang \textit{et al.} \cite{wang2017obtaining} used a small portion of high-quality instances to build models to classify the difficulty of unlabeled instances. Liu \textit{et al.} \cite{Liu2021ExploitingPA} used predicted labels to improve the performance of label aggregation. Their method captures the characteristics of workers and questions through neural networks, predicts the answers of different workers to the questions, and expands the label set to enhance the performance under sparse data. It is interesting that even transfer learning can be used to improve the truth inference \cite{Han2020CrowdsourcingWM,Xu2022Crowd}. These studies show that both supervised and unsupervised learning can be used to improve the truth inference.

In some early probabilistic graphical model-based inference algorithms \cite{welinder2010multidimensional,Yan2010ModelingAE,yan2011active,Yan2013LearningFM,bi2014learning,Zhao2015CrowdSelectionQP}, the instance features are retained in the models and participate in truth inference together with noisy labels. However, the role of the feature structures of instances in these inference models is still unclear. That is, how to select instance features and how to make the features contribute greatly to the improvement of the accuracy of truth inference requires further studies. In recent years, another new train of thought to utilize instance features in label aggregation resorts to deep learning. As early as 2016, Gaunt \textit{et al.} \cite{gaunt2016training} began to train deep neural networks with two building blocks, namely DeepAgg, for label aggregation. Yin \textit{et al.} \cite{yin2017aggregating} proposed label-aware autoencoders (LAA) to aggregate crowd wisdom. More sophisticated, Rodrigues and Pereira \cite{rodrigues2018deep} proposed a model CrowdLayer that trains deep neural networks to realize end-to-end learning from crowds (including label aggregation). Chen \textit{et al.} \cite{Chen2020StructuredPE} proposed SpeeLFC, which extends CrowdLayer with interpretable parameters and strengthens the correlation between workers and classes. GCN-Clean \cite{Iscen2020GraphCN} uses graph convolution networks (GCNs) to learn the relations between classes. The learned GCN model is used to clean wrong crowdsourced labels. Cao \textit{et al.} \cite{Cao2019MaxMIGAI} and Li \textit{et al.} \cite{Li2020CoupledViewDC} simultaneously aggregate the crowdsourced labels and learn an accurate classifier via multi-view learning. Yin \textit{et al.} \cite{Yin2020AggregatingCW} proposed a clustering-based label-aware autoencoder for label aggregation. The method uses clustering to aggregate instances with similar features, and constructs a deep generation process to infer the true labels. Li \textit{et al.} \cite{Li2021CrowdsourcingAW} proposed a fully Bayesian deep generative crowdsourcing model (BayesDGC), which combines the deep neural networks on automatic representation learning and the interpretable probabilistic structure encoding of probabilistic graphical models. It is worth noting that many deep learning models such as \cite{gaunt2016training,rodrigues2018deep,Chen2020StructuredPE,Iscen2020GraphCN} perform better when they have a small number of training examples with ground truth, which violates the agnostic characteristics of truth inference.

In summary, instance features potentially help improve the accuracy of truth inference, but how to develop more domain-independent methods requires further investigations.

\subsection{Perspective}
The objective of the data dimension is to provide high-quality data for knowledge model learning. The first direction to do so is that we can enrich the labels from different aspects. This derives the recent hot research on multi-label truth inference, whose core issue is to explore and exploit label correlations, which not only improves the inference accuracy but also reduces the number of labels required. Bragg \textit{et al.} \cite{bragg2013crowdsourcing} proposed multi-label naive Bayes (MLNB) model. For each label in the model, MLNB constructs a star graph with directed edges from that label to all other labels, which is used to calculate label correlations. Duan \textit{et al.} \cite{duan2014separate} studied how to extend the DS model in multi-label settings, proposing P-DS and ND-DS models. The P-DS model groups candidate labels into pairs, and then separately estimates the states of each pair, thereby preventing interference from uncorrelated labels. The ND-DS model depicts the conditional independence properties of the joint distribution over candidate labels as a Bayesian network and approximates the underlying joint distribution by the product of the conditional distributions of candidate labels. Zhang and Wu \cite{zhang2018multi} proposed a multi-class multi-label dependency (MCMLD) model. MCMLD introduces a mixture of multiple independently Multinoulli (or so-called Categorical) distributions to capture the correlation among the labels, together with a set of confusion matrices modeling the reliability of the workers. However, this model is very time-consuming. The variant of this model, MCMLD-OC \cite{Zhang2021MultiLabelTI}, limits each label to binary values, thereby reducing the complexity of the model. In addition to these generative methods, Tu \textit{et al.} \cite{Tu2020MultilabelCC} proposed a multi-label aggregation model MLCC based on joint matrix decomposition. MLCC decomposes the instance-label matrix into the product of two low rank matrices, and uses them to model the worker similarity and the correlation between labels. Because graph neural networks have strong correlation modeling capabilities, the use of graph neural networks in crowdsourced multi-label truth inference is a promising research direction.

However, only focusing on labels will limit the benefits of using crowdsourcing as a means of data collection to a narrow scope. Crowdsourcing should provide more variety of data for the upper-level model training. Since crowd workers can provide class labels, they can also provide descriptive data for instances themselves from various facets, namely Multi-Faceted Feature Description in Fig.~\ref{figArch}. Some attempts have been made in this direction. Deng \textit{et al.} \cite{deng2013fine} proposed an online game that reveals discriminative features of images, and the human annotated features are used to training classification models. Similar work can be found in \cite{zou2015crowdsourcing}. However, in these studies, human feature selection process does not introduce additional data. It is of great practical value for crowd workers to provide multi-faceted feature descriptions. For example, in a health record classification task, different doctors, nurses, and dieticians may provide different answers from their different concerns and perspectives. Therefore, in addition to the final class label, the reasons for their judgment must also be collected for model training. Currently, there are no studies focusing on integrating crowdsourced multi-faceted feature descriptions. In data collection and integration, we probably need the help of domain knowledge. External knowledge graph may serve as a reliable knowledge source. Therefore, how to introduce knowledge graph in crowdsourced data aggregation has become another topic. At least there are two ways that we can use knowledge graph to improve the label quality. In the data acquisition phase, the domain knowledge can be extracted and pushed to workers according to their current tasks so that they can obtain some hints to better complete the tasks. In the truth inference, domain knowledge can also be integrated into the true inference process to obtain better results.

Using topic models as a bridge to connect the conceptual inference model, multi-faceted feature integration model, and external knowledge graph is a potentially feasible solution. The topic model is a common way for knowledge representation, which can extract the most core knowledge concepts in the data and ignore those minor details. The probabilistic graphic representation of topic models makes it technically compatible with most crowdsourcing truth inference models, which can be uniformly integrated into larger graphical models. For each crowd worker, we can build a quality vector for the topics. This topic quality vector is also inferred from existing data, which describes the reliability of a worker on different topics at a fine-granular level.

\section{Robust and Complex Model Learning}
After the crowdsourced data are fused, we can use the collected data to build predictive models. Model learning is highly relevant to application domains. This paper only focuses on those domain-independent techniques.

\subsection{Predictive Model Learning}
Most of the current work focuses on supervised learning from crowdsourced labeled data. Compared with the truth inference in crowdsourcing that has already achieved fruitful results, the general-purpose predictive model learning research is still in its young stage.

\subsubsection{Standard Weakly Supervised Learning}
Weak supervision is defined as a particular supervised learning setting, where limited, or imprecise sources are used to provide supervision signals for labeling a large number of training data. Since the training instances have their integrated labels (though imperfect) after truth inference, it is straightforward to train learning models using any suitable learning algorithms, such as decision tree, SVM, etc. For example, in the early work \cite{sheng2008get}, a random forest is built after performing majority voting, and in \cite{raykar2010learning,bi2014learning}, logistic regression models are built after labels aggregated. These studies have demonstrated a basic fact that truth inference is an effective means to improve the quality of labels of training samples. Although imperfect, the noisy labeled training sets still can be used to train realistically available models. The performance of learned models not only depends on the quality of labels but also depends on the quality of instance features and learning algorithms. However, we should know that for some particular domains, label noises do deteriorate the learned models. This is the so-called \textit{two-stage} learning paradigm, i.e., inference \textit{plus} learning. There are also some other methods that directly build learning models from the raw data with repeated noisy labels. Kajino \textit{et al.} \cite{Kajino2012Convex,kajino2013clustering} proposed two methods, Personal Classifier and Clustered Personal Classifier, to learn logistic regression models with convex optimization. Both methods treat crowd workers as independent classifiers, each of which only uses the labels provided by a particular worker. All classifiers are modeled by a multi-task learning model with an objective function that can be globally optimized. Due to the limited effect of regression models in modeling high-dimensional sparse data, these methods can only be applied in a few fields. Donmez and Carbonell \cite{donmez2008proactive} proposed a Proactive learning method that does not include inference, but it merely works under the scenario that each instance is labeled by two workers. In \cite{sheng2011simple}, the author proposed a pairwise training strategy, where each instance has two weighted copies with negative and positive labels. We prefer two-stage methods for three reasons. First, we have not observed any extensive performance improvement for the bundled model (without inference). Moreover, when knowledge learning systems encounter problems, two-stage methods are easier to judge which part is out of order. The bundled model blurs the boundaries between inference and training. In addition, if we have plenty crowdsourced labeled data, the two-stage scheme will facilitate to build predictive models using a portion of data and use them to correct those incorrect labels \cite{zhang2018improving}.

\begin{figure*}
	\centering
	\includegraphics[width=6in]{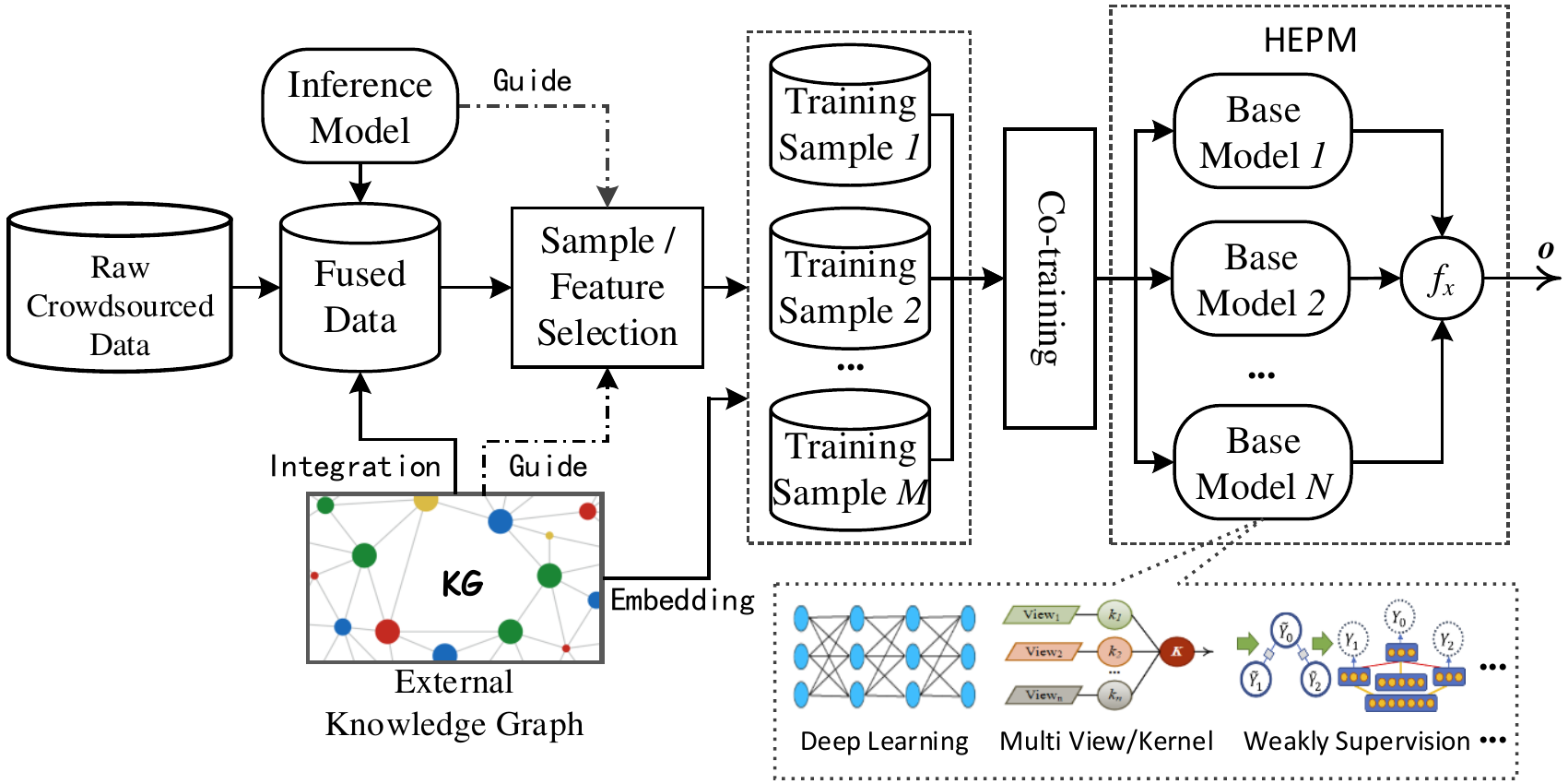}
	\caption{A blueprint to build robust and complex learning models. HEPM: Heterogeneous Ensemble Predictive Models.}
	\label{figLearning}
\end{figure*}

\subsubsection{Other Learning Settings}
Although using integrated labels to train models is a standard form, it may lose some information during inference. Noisy labels reflect the judgment of the workers to the instances, which may improve the generalization performance of learning models. To take advantage of full noisy labels, Sheng \cite{sheng2011simple} proposed five label utilization strategies for weakly supervised learning, which utilized the fact that some learning algorithms such as cost-sensitive decision trees \cite{Lomax2013ASO} and neural networks can accept weights for training instances. Thus, weights are generated for instances from the repeated labels, which are calculated using both frequency and the tail of a Beta distribution.

Some studies focused on learning methods in particular cases. Zhang \textit{et al.} proposed a method PLAT \cite{zhang2015imbalanced} that can automatically adjust the decision threshold between the inferred positive and negative instances, which can solve the imbalanced learning issue resulted from the biased labeling. This work showed that, biased labeling would exacerbate the imbalance of data distribution. Therefore, in the dual context of imbalanced underlying data and biased labeling, the objective of truth inference is not simply to achieve the maximization of accuracy but to maximize the imbalanced learning performance (for example, to maximize the AUC index in performance measure). Rodrigues and Pereira \cite{rodrigues2018deep} proposed a deep learning method CrowdLayer for crowdsourcing. Their method uses an EM algorithm to jointly learn the parameters of networks as well as the reliability of workers and then introduced a crowd layer that directly trains end-to-end deep neural networks from the noisy labels using back-propagation. This model adds a crowd layer after the traditional convolutional neural network (CNN). The crowd layer can be used for label aggregation during training. After the model training is completed, the crowd layer can be removed, and the remaining part is the standard CNN prediction model. Atarashi \textit{et al.} \cite{atarashi2018semi} addressed semi-supervised learning using deep neural networks. They presented a generative deep learning model, which leverages unlabeled data effectively by introducing latent features and data distribution. Shi \textit{et al.} \cite{Shi2020SemiSupervisedML} proposed a deep generative model for more complicated multi-label semi-supervised learning, which incorporates latent variables to describe the labeled/unlabeled data as well as the labeling process of crowdsourcing. Wang \textit{et al.} \cite{Wang2021DeepNL} believed that the inconsistency of crowdsourced labels not only stems from malicious workers or errors made by normal workers but also indicates semantic information (such as ambiguity or difficulty) of instances. Their method measures label inconsistency and assigns different weights to labels with different inconsistencies, which helps the training of neural networks and improves the learning performance. As the application of deep learning becomes more widespread, crowdsourcing learning will be more closely integrated with it.

\subsection{Perspective}
Due to the uncertainty of crowdsourcing, there are inevitably errors in the fused data. The target of the model dimension is to build complicated learning models that are robust to the errors. We have four types of data sources, i.e., conceptual annotation, multi-faceted feature description, instance features, and external knowledge graph, which allows the model dimension to adopt rich learning paradigms.

Fig.~\ref{figLearning} illustrates a blueprint for forecasting future technical development in this dimension. Usually, performing instance and feature selections on the dataset before model learning can effectively reduce noises, balance data distribution, and improve learning performance. The quantitative relationship between the various factors in statistical inference models for crowdsourcing, such as the reliability of workers, quality of topics, difficulty of instances, distribution of classes, quality of integrated labels, etc., can be established by probably approximately correct (PAC) learnable or statistical-query learnable theories \cite{lin2014re}. This quantitative relationship, together with the external knowledge graph, will provide some guidance information for the instance and feature selection processes.

Since both multi-source heterogeneous crowdsourced data and external knowledge graph describe the different facets of instance features, we can obtain a set of training datasets through the above instance and feature selection. From the perspective of the model training process, co-training has provided a potential solution to use different feature sets that provide different, complementary information about the instance. As a semi-supervised method, it learns a separate classifier for each view and predicts unlabeled or imperfect labeled data by iteratively constructing additional labeled training sets. From the perspective of decision making, ensemble learning can aggregate the outputs of multiple classifiers to form a more accurate prediction. Since the base models can be built with different learning algorithms from different views of features, heterogeneous ensemble predictive models can be created, which are more robust to the noises in the data according to the latest study \cite{zhang2018ensemble}. Moreover, in their ensemble model, each instance is duplicated with different weights according to the distribution and class memberships of its multiple noisy labels and the final classifier is obtained from the aggregation of multiple base learners using the maximum a posteriori probability estimate, which shows a smaller upper boundary of error rate than that of the voting method. This ensemble model is isomorphic, where the same learning algorithm runs on the training sets extracted from the same data pool. If training sets from different facets could be used, heterogeneous ensemble learning may achieve better performance.

From the perspective of specific learning algorithms, there are also some research topics that have never been well studied. Although some recent studies \cite{rodrigues2018deep,atarashi2018semi,Shi2020SemiSupervisedML} push the crowdsourcing learning into a deep learning stage, how to optimize the deep learning model to cope with crowdsourcing noises has not been studied as well as embedding external knowledge graph to enhance deep learning models \cite{wang2017knowledge}. We believe that crowdsourcing learning based on graph neural networks \cite{Wu2019ACS} is a promising research topic. For example, we can let the network nodes represent the crowdsourced workers, and let the weight of the network edge represent the similarity of the workers, so as to obtain the adjacency matrix. Then, we define the degree matrix of the worker network, that is, the degree matrix only has values on the diagonal, indicating the number of workers adjacent to a node. All the noise labels of all workers on a certain sample form a matrix. In this way,  three types of required matrices of the graph convolutional neural network (GCN) \cite{Kipf2017Semi} are obtained. GCN has two convolution operations, and the integrated labels of the samples can be obtained by averaging all rows of the output label matrix of the second convolution operation. After obtaining the integrated labels, we connect GCN with a traditional CNN to train predictive models. On GCN, the idea of K-nearest neighbors can be used to complete missing labels. We can also use the predicted labels obtained from the CNN fully connected layer and the integrated label obtained from GCN to calculate cross-entropy loss to train the parameters on the convolutional layers, max-pooling layers, and fully connected layers of CNN.

The multi-view and multi-kernel learning is another interesting point, which can build learning models directly on different feature subspaces or heterogeneous feature spaces. Zhou and He \cite{Zhou2017ARA} made the first attempt on this topic, however, we have not seen much further  research. Some weakly supervised learning techniques, such as graph matching in semi-supervised learning, self-taught learning \cite{fang2012self}, data programming \cite{ratner2016data}, positive-unlabeled learning \cite{hsieh2015pu}, noise filtering and correction \cite{Zou2021UnsupervisedEL}, etc., can be used to build noise-robust models.

\section{Learning Process and Strategies}
In real-world applications, the ideal state is that the knowledge learning system maintains the ability of continuous learning at low cost and is readily available. Therefore, the focus of the systemic dimension is the optimization of the knowledge learning system, which includes a trade-off between cost and performance, dynamic modeling of the system state, and the guarantee of the high availability of the system. All these functionalities need the support of a knowledge base, which, although necessary, belongs to the research area of data management \cite{li2016crowdsourced}, which is out of the scope of this paper. During the past years, active learning with crowdsourcing attracted much attention from researchers because it can significantly reduce the cost of crowdsourced annotation while maintaining high learning performance. Active learning is defined as a special learning paradigm, where a learning algorithm can interactively query a user (or some other information source, i.e., a teacher or an oracle) to label new data points with the desired outputs. There is a natural connection between crowdsourcing learning and active learning. Since crowdsourcing labeling can significantly reduce labeling costs compared to expert labeling, the pursuit of optimization of labeling costs naturally becomes one of the goals of crowdsourcing learning. The realization of this goal requires the help of active learning. Therefore, this section puts forward our viewpoints on future technical development after reviewing the progress in active learning.

\begin{figure*}
	\centering
	\includegraphics[width=5in]{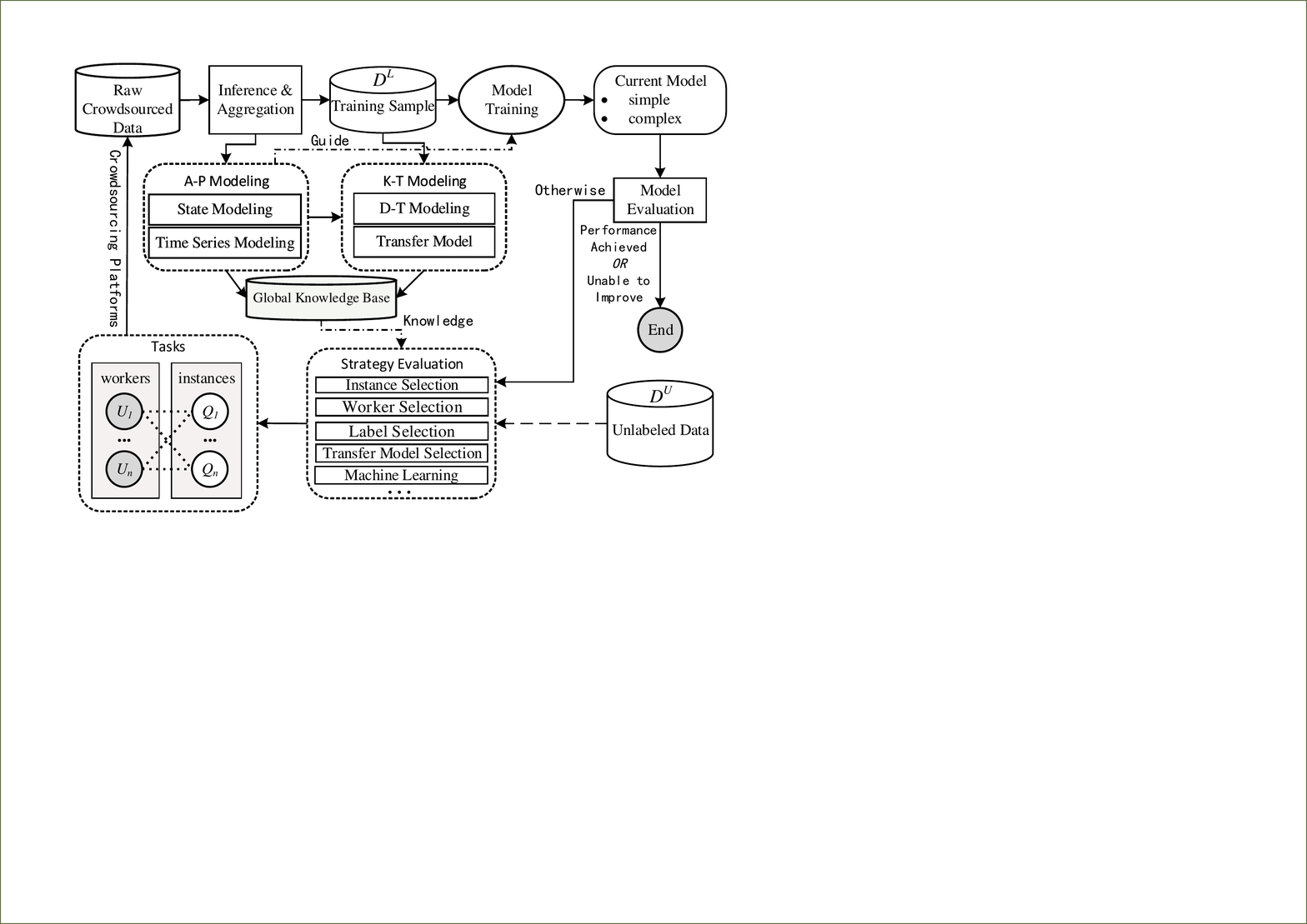}
	\caption{A blueprint of the technical development trend and their relations in the systemic dimension. A-P: Annotation Process; K-T: Knowledge Transfer; D-T: Domain Topic.}
	\label{figSystem}
\end{figure*}

\subsection{Active Learning and its Strategies}
Active learning optimizes the learning process through the design of learning strategies. The early work \cite{sheng2008get} first proposed three uncertainty-based instance selection strategies for crowdsourcing which consider the noisy label distributions on instances, the prediction of current models, and both of them, respectively. In this work, the label uncertainty is calculated from the distribution of two types of labels in the sample's noisy crowdsourced label set, which reflects the inconsistency of crowdsourced workers' judgments on sample types. Model uncertainty is calculated based on the probability that the sample belongs to each class calculated by the current model, that is, the greater the entropy, the more uncertain. The hybrid uncertainty is the geometric mean of the label uncertainty and model uncertainty, which has the best overall performance. However, it can be seen from the experiments of this work that label uncertainty is very sensitive to worker errors, and a small number of labeling errors will cause the fluctuation of model's performance. This will increase the difficulty of judging the convergence of the model in practice. Zhang \textit{et al.} \cite{zhang2015active} extended these strategies to a biased labeling scenario by considering the level of labeling bias obtained from their PLAT algorithm, which partially solves the imbalance issue caused by biased labeling. Yan \textit{et al.} \cite{yan2011active} believed that selecting crowd workers is also necessary to improve the quality of labels and should be treated as a strategy in active learning. They designed a strategy that picks the workers that are most beneficial to the performance improvement of the current learning model during each iteration of active learning. However, this work did not answer the question that if the optimal worker is unavailable (for example, temporarily withdrawing from the task), choosing a sub-optimal worker will bring what kind of impact on the model. Long \textit{et al.} \cite{long2013active} proposed the Bayesian active learning that introduces a sorting strategy based on information entropy when selecting instances and workers. Even the more complex multi-label active learning follows a similar research idea, but more correlation modeling is added by evaluating the instance-label pairs \cite{bragg2013crowdsourcing,Li2019MultiLabelLF}. As a mode of data acquisition, active learning paradigm also can be used for label aggregation \cite{Yu2020ActiveMC}. 

In addition to the above standard forms of active learning, some studies have designed more diverse active learning strategies. Self-taught active learning \cite{fang2012self} first classifies crowd workers into two categories, i.e., weak workers and reliable workers, and then occasionally selects weak workers but uses the labels provided by reliable workers to compensate for their working outputs. This strategy not only learns reliable knowledge but also provides opportunities to those weak workers and help them improve their capability, which prevents the system from being unavailable after the reliable workers leave. However, the judgment of who is a reliable worker in this work seems to be derived from prior knowledge. In addition, the work did not consider the strategy of label expansion for an unreliable worker, that is, which reliable workers will be used for the expansion of the unreliable worker. Lin \textit{et al.} \cite{lin2016re} proposed a concept of re-active learning that introduces two sampling strategies--uncertainty sampling and impact sampling. It makes a trade-off between acquiring more labels to lower the noise levels and enlarging the size of the train set with more instances. It is a common sense that although some workers exhibit a low quality, they can still provide correct answers to some specific instances. This work was very impressive, because most other algorithms are guided by the designed active learning strategy to automatically decide whether to expand the crowdsourced label set of a sample or to select a new sample. This work explicitly considered this issue before the active learning strategy was designed. Unfortunately, we did not find more in-depth investigations of this issue. According to this fact, Huang \textit{et al.} \cite{huang2017cost} proposed an active strategy that evaluates the cost-effectiveness of instance-worker pairs. The strategy selects an instance that is beneficial for the performance improvement of the current model and a worker that has a high probability of providing a correct answer to this instance at a relatively low cost.

Many active learning strategies are heuristic. They are intuitively rational, but it is difficult to conduct a theoretical analysis. It is impossible to know the performance boundaries of these strategies. Therefore, theoretical research in this direction needs to be strengthened in the future.

\subsection{Perspective}
Due to the uncertainty of crowdsourcing, the functionalities of the systemic dimension are extremely critical to crowdsourced knowledge learning systems. There are several examples. Sometimes our active learning strategy does choose the most appropriate worker to perform the task. However, when the task is pushed to the worker, the worker has left the system, or the correct rate of the worker begins to decline due to long hours of work, which fails the active learning and results in the unstableness of the knowledge learning process. In some other cases, a requester publishes a set of new human intelligence tasks, however, workers in the system are not familiar with this application domain, resulting in a low quality of answers and bad learned models. All these issues need to be addressed with new techniques in the systemic dimension.

Fig.~\ref{figSystem} illustrates a blueprint of the technical development trend and their relations. The entire technical roadmap adds a variety of modeling processes, strategy design, and a global knowledge base into the crowdsourcing active learning framework. To ensure the stability of the knowledge learning process, we need to dynamically model the crowdsourcing annotation process, including the system state modeling and time series modeling. There has been some work addressing the dynamical modeling of crowdsourced annotation. Rodrigues \textit{et al.} \cite{rodrigues2014gaussian} introduced a Gaussian process classification to model multiple annotators with different levels of expertise. Jung \textit{et al.} \cite{jung2014predicting} explored temporal behavioral patterns of underlying crowd work and proposed a time-series label predictive model to capture past worker behaviors. Modeling the state and time series of a crowdsourcing system relies on the results of the statistical inference and aggregation of multi-source data. For example, we may use Hidden Markov Models (HMM), State Space Models, or Latent Autoregressive Models to model the changes of the reliability of workers over times. Compared with statistical inference, dynamic time series modeling can reduce the influence of early historical data on current state prediction. Accurate time series modeling will increase the effectiveness of the instance and worker selection.

\begin{table*}[]\centering
	\caption{Accuracy of Seven Truth Inference Method on Eight Real-World Datasets (in Percentage)}
	\label{tab:Perf}
	\begin{tabular}{c|ccccccc}\hline
		Dataset     & MV      & DS    & Spectral DS & ZenCrowd & BCC   & GTIC & CrowdLayer \\ \hline
		Fej2013     & 90.28   & 87.67 & 88.62       &   90.10  & 89.02 & \textbf{90.76} & 88.20           \\
		Trec2010    & 44.20   & 50.27 & 47.59       &   30.98  & 46.88 & 45.48 & \textbf{51.02}           \\
	    AdultContent& 75.68   & 73.57 & 66.06       &   75.98  & 76.08 & \textbf{77.78} & 74.30           \\
	    Valence     & 36.00   & 40.00 & 45.00       &   32.00  & 40.00 & \textbf{54.00} & 48.88           \\
	    Duck        & 68.80   & 60.80 & 72.20       &   58.80  & 63.38 & 74.20 & \textbf{75.04}          \\
	    WordSim     & \textbf{90.00}   & \textbf{90.00} & 89.00       &   86.70  & 87.20 & 86.70 & 88.40           \\
	    Aircrowd    & 81.10   & 91.92 & \textbf{91.73}       &   81.45  & 87.42      & 81.96 & 90.02           \\
	    Leaves      & 64.16   & \textbf{64.28} & 58.62       &   59.70  & 60.98      & 61.46 & 63.80    \\ \hline       
	\end{tabular}
\end{table*}

Another interesting research direction is knowledge transfer modeling in the systemic dimension. Knowledge transfer can improve the usability and stability of a knowledge learning system. Particularly, when the system is in its cold-start stage, we can use the knowledge of related domains that has been learned before to improve the quality of the outputs of crowd workers. Researchers have already noticed the effect of transfer learning for the crowdsourcing inference. Mo \textit{et al.} \cite{mo2013cross} first introduced transfer learning into crowdsourcing and proposed the cross-task crowdsourcing that could share the knowledge across different domains and solve knowledge sparsity of a particular domain. Fang \textit{et al.} \cite{fang2013knowledge,fang2014active} proposed an active learning framework with knowledge transfer, where workers' expertise is modeled from the historical annotation in a source domain and used in a target domain in the instance and worker selection. Zhao \textit{et al.} \cite{zhao2013transfer} transferred the knowledge from categorized \textit{Yahoo! Answers} datasets for learning user expertise in the tasks on Twitter. The learning strategy of the knowledge transfer model solves the \textit{cold-start} problem in the process of new topic expansion in crowdsourced annotation. We need to embed the knowledge transfer process in active learning and start this process under certain conditions. Knowledge transfer is a way of knowledge sharing between the source domain and the target domain. It can use the model of the source domain to determine or predict the attributes of samples in the target domain. The source domain is composed of samples that have been well labeled and formed a good prediction model for the specific field. The target domain consists of samples that are ready to be crowdsourced for annotation (description). In the iterative process of active learning, the existing sample selection strategy is first used to select samples that need to be crowdsourced labeled, and then we evaluate whether these samples need knowledge transfer. We evaluate the similarity between each topic in the source domain and the topics of these samples. If the topic of the samples is different from the topics of the source domain, the knowledge transfer process needs to be initiated, otherwise, the original active learning strategy will be used. In addition, for those already labeled samples, how to use them to update the source domain model also needs further consideration.

We have noticed that, compared with traditional active learning, active learning from crowds can hardly maintain a smooth rising learning curve. This may be partially caused by the reason that workers probably make mistakes repeatedly due to the lack of expertise or the exhaustion of repetitive tasks. Thus, machine teaching \cite{zhu2015machine}, the inverse problem of machine learning, can be embedded in the active learning procedure, where the selected teaching examples are pushed onto the workers when they perform annotation tasks. The latest study \cite{Zhang2020InteractiveLW} attempted to enhance the ability of workers by the Generalized Context model in cognitive psychology. Therefore, we need to introduce the concept of interactive learning, which can feed back knowledge to crowdsourcing workers in the learning process to maintain and improve their working ability. The core part of interactive learning is the online machine teaching process. The difficulty lies in the generation of teaching cases. After selecting the samples to be labeled in the next round, interactive learning needs to generate some teaching cases to help the crowdsourced workers complete the tasks with high quality. The number of these teaching cases cannot be too many, and at least the teaching needs to be carried out from both positive and negative aspects. Too many teaching cases will not only increase the learning burden of workers, but also distract workers' attention when performing tasks. Providing positive and negative teaching cases simultaneously can enable workers to quickly grasp concepts through comparison. Therefore, we need to find those samples that are most similar to the samples to be labeled and those that look similar but have different categories in the knowledge base as teaching cases. In addition, we have to consider how to update the teaching knowledge base.

To sum up, under this framework, active learning strategies include the instance, worker, transfer-model selections, and even machine teaching methods. The transfer-model selection strategy aims to determine whether the current task requires a transfer model and which transfer model to use. Once selected, it should be considered in both instance and worker selections. Finally, the human intelligence tasks adopt a two-part graph structure. When workers accept tasks, they also receive some recommended references which may improve their skills by machine teaching, finally achieving \textit{positive feedback} knowledge learning.

\section{Research Tools}
Open-source tools including datasets for public acquisition is a trend in recent data-driven research. There have been some open-source research tools for crowdsourcing learning. Nguyen \textit{et al.} \cite{nguyen2013batc} proposed a visual tool BATC for label aggregation research, which implements
several truth inference algorithms MV, DS, RY, KOS, and GLAD but uses synthetic datasets. The advantage of this tool is that it provides a easy-to-use graphical user interface for both evaluation and simulation. Sheshadri and Lease \cite{sheshadri2013square} proposed another tool SQUARE for truth inference. It does not provide a graphic user interface but provides a set of APIs that facilitate users to integrate their functions. SQUARE implements and integrates the algorithms
MV, DS, RY, GLAD, and ZenCrowd. Moreover, it collects ten real-world crowdsourcing data sets. In \cite{zheng2017truth}, the authors provided dozens of implementations of truth inference algorithms together with several real-world datasets. Zhang \textit{et al.} \cite{zhang2015ceka} proposed a tool CEKA, which involves model learning processes as well as a large number of ground truth inference algorithms. It follows the object-oriented design and is fully compatible with a well-known machine learning tool WEKA \cite{hall2009weka}. CEKA has a more open architecture, which makes it easy to integrate new algorithms in the future. Table~\ref{tab:Perf} shows an evaluation example of CEKA. We evaluated seven different truth inference methods implemented and integrated in CEKA on eight real-world crowdsourcing annotation datasets. The experimental results show that GTIC and CrowdLayer have good performance in general. Some traditional inference methods such as MV and DS are not necessarily bad. Venanzi \textit{et al.} \cite{Venanzi2015TheAA} presented an open-source toolkit that allows the easy comparison of the performance of active crowdsourcing learning methods over a series of datasets. Users can construct new strategies by combining aggregation models, task selection methods, and worker selection methods. However, this tool only implemented a few existing algorithms and collected a small number of real-world crowdsourcing datasets.

In the future, it is worth studying how to integrate open source research tools with real crowdsourcing platforms to support the entire knowledge learning process. The development of open source tools still faces some challenges. First, the tools need to simulate the complex behaviors of crowdsourced workers to support in-depth analysis of the performance of different algorithms in different environments. As we have seen in many studies, we have quite a few truth inference methods, actually, we still do not have a very clear understanding of the applicability of these algorithms. The behaviors of these crowdsourced workers are often not independent, and they may influence each other in large tasks. Second, the tools need to have interfaces that support mainstream crowdsourcing platforms. These interfaces make it easy for researchers to connect to the real crowdsourcing platform, and make the self-made back-end systems work together with the crowdsourcing platform. Third, the tools need to support mainstream programming languages and frameworks. For example, in the field of deep learning, researchers are used to using Python language and working with PyTorch or TensorFlow deep learning frameworks. In addition, in the field of crowdsourcing learning, there is an urgent need for comprehensive comparison of deep learning methods with traditional learning methods. Finally, visualization is a very necessary but difficult traditional issue.

\section{Conclusion}
Knowledge learning with crowdsourcing has launched an enormous picture for the researchers in AI-related disciplines. This paper summarized the progress in the field from a systematic perspective, including three dimensions of techniques to promote the performance of knowledge learning. More importantly, according to our many years of research experience, this paper comprehensively discusses the future research directions in this field from the perspective of knowledge learning for the first time. In the data dimension, we emphasize to maximize the heterogeneous data collection ability of crowdsourcing and aggregate various types of crowdsourcing data. In the model dimension, we emphasize the use of different learning paradigms, training methods, and the characteristics of the learning model to make full use of data. In the systemic dimension, we emphasize the optimization of model training costs, system reliability, and human-machine collaboration.


%

\section*{Acknowledgment}
The author thanks all researchers in the relevant fields for their contributions to the technical progress as well as the editors and reviewers for their efforts in the publishing of the article in the IEEE/CAA Journal of Automatica Sinica.

\ifCLASSOPTIONcaptionsoff
  \newpage
\fi



\bibliographystyle{IEEEtran}
\bibliography{manuscript}
\end{document}